\documentclass{article}

\usepackage{microtype}
\usepackage{graphicx}
\usepackage{subfigure}
\usepackage{booktabs}
\usepackage{amsmath}
\usepackage{algorithm}
\usepackage{algorithmic}
\usepackage{amsfonts}
\usepackage{hyperref}

\usepackage[accepted]{icml2021}

\icmltitlerunning{Statistical Undersampling with Mutual Information and Support Points}

\begin{document}

\twocolumn[
\icmltitle{Statistical Undersampling with Mutual Information and Support Points}



\icmlsetsymbol{equal}{*}

\begin{icmlauthorlist}
\icmlauthor{Alex Mak}{equal,uamss}
\icmlauthor{Shubham Sahoo}{equal,uamss}
\icmlauthor{Shivani Pandey}{uamss}
\icmlauthor{Yidan Yue}{uamss}
\icmlauthor{Linglong Kong}{uamss}
\end{icmlauthorlist}

\icmlaffiliation{uamss}{Department of Mathematical and Statistical Sciences, University of Alberta, Edmonton, Canada}
% \icmlaffiliation{goo}{Googol ShallowMind, New London, Michigan, USA}
% \icmlaffiliation{ed}{School of Computation, University of Edenborrow, Edenborrow, United Kingdom}

\icmlcorrespondingauthor{Alex Mak}{amak@ualberta.ca}
\icmlcorrespondingauthor{Shubham Sahoo}{spsahoo@ualberta.ca}

\icmlkeywords{Class Imbalance, Support Points, Mutual Information}

\vskip 0.3in
]



\printAffiliationsAndNotice{\icmlEqualContribution} % otherwise use the standard text.

\begin{abstract}

Class imbalance and distributional differences in large datasets present significant challenges for classification tasks machine learning, often leading to biased models and poor predictive performance for minority classes. This work introduces two novel undersampling approaches: mutual information-based stratified simple random sampling and support points optimization. These methods prioritize representative data selection, effectively minimizing information loss. Empirical results across multiple classification tasks demonstrate that our methods outperform traditional undersampling techniques, achieving higher balanced classification accuracy. These findings highlight the potential of combining statistical concepts with machine learning to address class imbalance in practical applications.

\end{abstract}

% Part 1
% --------------------------------
\section{Introduction}
\label{submission}

Class imbalance is a significant challenge in machine learning, where datasets have skewed class distributions with one or more classes dominating others. This imbalance leads to poor performance in identifying minority class instances, which are often crucial in applications such as fraud detection, medical diagnosis, and fault detection \cite{Ganganwar, Devi}. Undersampling is a widely used data-level solution that balances the dataset by reducing the dominance of majority class instances.

Undersampling, while effective in addressing class imbalance, has limitations, such as the potential loss of critical information when majority class data is removed and the difficulty in handling class overlap near decision boundaries \cite{Devi,Ganganwar}. As a result, this report proposes advanced approaches leveraging support points and mutual information (MI) techniques to look for improved undersampling solutions.

Particularly, the forward steps in this research involve applying support points and MI-based stratification techniques to optimize data selection processes. By focusing on informative and representative data subsets, the proposed approach aims to enhance classifier performance, particularly for minority class recognition, while maintaining computational efficiency and addressing the limitations of existing undersampling methods.

\section{Related Work}

\subsection{Mutual Information}

In terms of academic literature on applying mutual information in data sampling and data resampling, there is not a lot any relevant work in this area as most work centers around researching more accurate mutual information estimators. 

Nevertheless, Shang et al. proposed an approach of maximizing mutual information for dataset distillation, by synthesizing a small dataset whose predictive performance is comparable to the full dataset using the same machine learning model \cite{NEURIPS2023_24d36eee}. Specifically, Their research designed an optimizable objective in a contrastive learning framework to update the synthetic dataset which could be used in deep learning tasks \cite{NEURIPS2023_24d36eee}.

We were inspired by their research approaches, and we were seeking whether maximizing mutual information can be a research approach in the context of data undersampling since we found out the concept of data distillation is similar to data undersampling, where they share the common objective of preserving as much information as possible in a dataset's subset. 

\subsection{Support Points}
The use of \textit{support points} as an advanced sampling method was first introduced by Simon Mak and V. Roshan Joseph (2018) as a technique to generate representative subsets of data by minimizing the energy distance \cite{Mak-Support-Points}. Their work primarily focused on providing compact samples that preserve the statistical properties of continuous distributions, which has been widely applied in uncertainty quantification and Bayesian computation. Despite its success, the use of support points to address class imbalance in machine learning has remained unexplored. Our research extends the application of support points by adapting them to downsample majority-class data in imbalanced datasets. By selecting statistically representative subsets, our method preserves data integrity and avoids information loss commonly seen in random undersampling.

In contrast, Reshma C. Bhagat and Sachin S. Patil (2015) proposed an enhanced SMOTE algorithm for classifying imbalanced big data using Random Forest. Their method decomposed multi-class imbalanced datasets into binary subsets, where SMOTE was applied to generate synthetic samples, achieving balanced data. While effective, SMOTE-based techniques can introduce synthetic samples that may not accurately represent the true data distribution, often lacking interpretability and adding computational complexity. Our approach tackles this limitation by focusing on downsampling instead of oversampling, using support points to ensure that selected samples remain statistically consistent with the original data while improving model performance on minority classes.

\section{Problem}

Altogether, imbalanced datasets present a significant challenge in machine learning, it impacted machine learning models to develop a bias toward the majority class, resulting in poor predictive performance especially for the minority class. This bias can lead to suboptimal decisions, particularly in high-stakes fields such as healthcare, fraud detection, and marketing where detecting rare anomalies is crucial.

Traditional undersampling methods attempt to address this issue by selectively removing data points from different criteria. However many failed to adequately preserve class distributions, thereby losing an excessive amount of information and representativeness during the undersampling process. 

As a result, we proposed two alternative approaches to undersample the majority class. The first approach centers around using mutual information and stratified simple random sampling (SRS); meanwhile the second approach emphasizes on the application of support points. 
We proposed and conducted this research with the aim of offering an improved undersampling process that can address class imbalance and improve model performance on classification task by minimizing the impact of representative and information loss.

Our proposed research can be summarized to the following research question:
\textit{Can stratified simple random sampling (SRS), when combined with mutual information or support points provide a more effective undersampling solution compared to other techniques for addressing class imbalance?}

% --------------------------------

% Part 4
% --------------------------------
\section{Methodologies}

\subsection{Mutual Information}
\textbf{Overview and Motivation:} \\
Mutual Information is the amount of information obtained from one random variable by observing another random variable \cite{NEURIPS2023_24d36eee}. It can also be interpreted as reducing uncertainty in a random variable given the knowledge of another \cite{NEURIPS2023_24d36eee}. 

By stratifying data based on mutual information, similar data points will be grouped together. This allows the subsequent stratified simple random sampling (SRS) process to avoid sampling too many similar data, which allows retaining the sampled data's representativeness on the original data. 

\subsubsection{Mathematical Formulation}
According to Shang et al., the mathematical formulation of mutual information is straightforward \cite{NEURIPS2023_24d36eee}.

% Adjust space above and below the equation
\setlength{\abovedisplayskip}{3pt}
\setlength{\belowdisplayskip}{3pt}

\[
I(X, Y) = \sum_{x, y} P_{XY}(x, y) \log \frac{P_{XY}(x, y)}{P_X(x) P_Y(y)}
\]

where \(P_{XY}(x, y)\) is the joint distribution between x and y, \(P_{X}(x)=\sum_{y} P_{XY}(x, y)\) and \(P_{Y}(y)=\sum_{x} P_{XY}(x, y)\) are the maginal distributions of \( X\) and \( Y\) respectively. 

Based on the measure above, we performed an element-wise computation to calculate the mutual information value for each data point in a dataset, then grouped them through K-Means clustering in order to mimic the stratifying process in stratified SRS.

\subsubsection{Algorithmic Procedure}

The mutual information values in each data point in a dataset can be obtained with the following method:

\begin{enumerate}
    \item \textbf{Initialization:} Understanding there are \(X\) rows and \(Y\) columns in a data set,
    \item \textbf{Compute MI:} Calculate MI between each \(x\) and \(y\) pair.
    \item \textbf{Storing Values:} storing the MI values in a matrix, only calculate the top half triangle of a MI matrix since \(I(x, y)\)=\(I(y, x)\).
    
\end{enumerate}

\subsubsection{Assumptions}
According to Fernandes and Glorr, there are 2 empirical assumptions about mutual information\cite{Fernandes2010}.
\begin{enumerate}
    \item Joint probability can be estimated using observed frequencies \(\ n_{xy}\).
    \((P_{xy}\approx \frac{n_{xy}}{n})\), where \(n = \sum_{x}\sum_{y}n_{xy}\)
    \item The marginal probabilities can be defined by summing over the joint probabilities \((p_{i+} = \sum_j p_{ij} \, \text{and} \, p_{+j} = \sum_i p_{ij})\).
    
\end{enumerate}

\subsubsection{Complexity Analysis}

In our study, since we utilized a brute-force approach to compute mutual information where each pair of datapoints' mutual information is calculated individually. According to Ferdosi et al., a dataset \(n\) requires the calculation of mutual information over \(\binom{n}{2}\) times \cite{pmlr-v108-ferdosi20a}. Even though the MI matrix is symmetric and we intend to compute the top triangular matrix, the time complexity of computing mutual information is still \(O(n^2)\).

\subsection{Support Points Approach}

\textbf{Overview and Motivation:} \\
Support points represent a principled method for selecting a small, representative subset of samples from a larger dataset while preserving essential distributional characteristics. Instead of relying on arbitrary or random sampling, the support points approach seeks to minimize a well-defined statistical distance metric between the empirical distribution of the chosen subset and that of the original dataset. By doing so, it addresses class imbalance through an informed undersampling strategy that retains core structural information about the majority class.

The key intuition is that by minimizing a suitable distance measure—one sensitive to distributional differences—the reduced dataset (the support points) will capture the variability and structure of the original data more faithfully than naive subsampling. This is particularly important in imbalanced classification, where over-simplification of the majority distribution can distort decision boundaries and degrade minority class detection.

% --------------------------------
\subsubsection{Mathematical Background and Objective Function}

\textbf{Energy Distance:} \\
We adopt the \textit{energy distance} as the metric of discrepancy between distributions. Given two probability distributions \( F \) and \( G \) on \( \mathbb{R}^d \), the energy distance \( E(F, G) \) is defined as \cite{Mak-Support-Points}:

% Adjust space above and below the equation
\setlength{\abovedisplayskip}{3pt}
\setlength{\belowdisplayskip}{3pt}

\[
E(F, G) = 2\mathbb{E}\|X - Y\| - \mathbb{E}\|X - X'\| - \mathbb{E}\|Y - Y'\|.
\]

where \( X, X' \overset{\text{i.i.d}}{\sim} F \) and \( Y, Y' \overset{\text{i.i.d}}{\sim} G \). This metric is zero if and only if \( F = G \), making it a strong candidate for distributional matching.

\textbf{Empirical Approximation:} \\
In practice, we approximate distributions using finite samples. Let \( X = \{x_1, \dots, x_N\} \) be a dataset drawn i.i.d. from \( F \), and let \( Z = \{z_1, \dots, z_m\} \) (where \( m \ll N \)) represent the reduced set approximating \( F \). The empirical energy distance \( \hat{E}(X, Z) \) is given by \cite{Mak-Support-Points}:
\begin{equation}
\begin{split}
    \hat{E}(X, Z) &= \frac{2}{Nm} \sum_{i=1}^N \sum_{j=1}^m \|x_i - z_j\| 
    - \frac{1}{N^2} \sum_{i, i' = 1}^N \|x_i - x_{i'}\| \\
    &\quad - \frac{1}{m^2} \sum_{j, j' = 1}^m \|z_j - z_{j'}\|.
\end{split}
\end{equation}

Minimizing \( \hat{E}(X, Z) \) balances two effects:
\begin{enumerate}
    \item Bringing \( Z \) points close to the points in \( X \),
    \item Ensuring that the points in \( Z \) do not collapse into a single cluster, by encouraging spread among themselves.
\end{enumerate}

We aim to solve:
\begin{equation}
\min_{Z} \hat{E}(X, Z),
\end{equation}
where \( Z \) represents the support points.

% --------------------------------
\subsubsection{Algorithmic Procedure}

The support points subset \( Z \) is obtained by minimizing \( \hat{E}(X, Z) \) using gradient-based optimization. The algorithm proceeds as follows:

\begin{enumerate}
    \item \textbf{Initialize:} Randomly sample \( m \) points \( Z \) from the majority class \( X \).
    \item \textbf{Compute Gradients:} Compute the gradient of \( \hat{E}(X, Z) \) with respect to each \( z_j \). The gradient for each point \( z_j \) is given by:
    \begin{equation}
        \frac{\partial \hat{E}}{\partial z_j} = \frac{2}{Nm} \sum_{i=1}^N \frac{z_j - x_i}{\| z_j - x_i \|} - \frac{2}{m^2} \sum_{j' \neq j} \frac{z_j - z_{j'}}{\| z_j - z_{j'} \|}.
    \end{equation}
    \item \textbf{Update Points:} Move each \( z_j \) against the gradient direction using a learning rate \( \eta \):
    \begin{equation}
        z_j \leftarrow z_j - \eta \frac{\partial \hat{E}}{\partial z_j}.
    \end{equation}
    \item \textbf{Iterate:} Repeat the process until convergence, or for a fixed number of iterations.
    \item After convergence, the optimized \( Z \) is mapped to the nearest neighbors in \( X \), yielding \( Z_{\text{nearest}} \).
\end{enumerate}

% --------------------------------
\subsubsection{Assumptions}
\begin{itemize}
    \item The dataset has finite first moments, ensuring well-defined computations of energy distance.
    \item The reduced subset \( Z \) should adequately represent the distributional characteristics of the original dataset \( X \).
    \item The clustering-based subsampling step assumes that the chosen subset retains key modes and variability of the original data.
\end{itemize}

% --------------------------------
\subsubsection{Complexity Analysis}
Let \( N' \) be the size of \( X_{\text{subset}} \) and \( m \) the number of support points. Each iteration involves:
\begin{itemize}
    \item \textbf{Distance Computations:} \( O(N' m d + m^2 d) \),
    \item \textbf{Gradient Updates:} Dominated by the same distance computations.
\end{itemize}
For \( T \) iterations, the total complexity is \( O(T (N' m d + m^2 d)) \).

% --------------------------------

% Part 5
% --------------------------------
\section{Experiment Design}

To carry out our proposed research, we decided to conduct an experimental study. The experimental study we plan to conduct emphasizes on conducting different methodologies including mutual information and support point to undersample the majority class data in an imbalance dataset. To evaluate our proposed research, we fitted the resampled data to a classifier to perform classification and observe its impact.

To provide a more comprehensive review of our research, our study also included a baseline experiment for direct comparison purposes. Specifically, the baseline experiment is conducted by performing random undersampling to undersample the entire dataset. This allows us to observe our research's impact more precisely. In addition, to retain the reproducibility of our proposed research, our experiment has set a fixed random state number whenever possible.

\subsection{Experiment Steps}
Our experiment consists of the following steps: 
\begin{enumerate}
    \item \textbf{Data Collection and Preprocessing} \\
    First, an imbalanced data is loaded and processed. Data preprocessing is needed to ensure the data is ready for the other steps in the experiment. Specifically, our data preprocessing work emphasizes on data cleaning and data transformation. Data cleaning removes null and duplicate entries in our data because they serve no impact to the data itself. For missing values, they are often imputed with the feature’s mean if the feature holds continuous values, otherwise it will be replaced with the feature’s mode. In addition, data transformation is also conducted to encode any categorical features to have numerical data type. In addition, not all the data preprocessing methods mentioned above were used to process the imbalance data as we attempted to minimize our changes on any datasets, and keep them as similar as it was first loaded.  
    \item \textbf{Undersampling through Stratified SRS} \\
    After the data is processed, it is ready to be undersampled. As mentioned above, we have 2 main areas: mutual information and support points. We will apply one of the 2 approaches of our sampling work at a time. Their details will be further explain in the next section.
    \item \textbf{Evaluation through Classification} \\
    To evaluate the impact of our undersampling research, the resampled data is fitted to a classification problem. Specifically, the resampled data will be spitted in 80-20 training-testing split to train and evaluate the classifier respectively. The classifier selected for this experimental study are logistic regression, random forest, XgBoost (XGBoost), and support vector machine (SVC) without any tuning. The classifier is first trained with the training data, then makes predictions on the testing data. The made predictions are evaluated with an array of metrics, including balanced accuracy,  precision, recall, and F1-score. These metrics not only capture undersampled data’s impact on the classifier’s performance, specific class performance will be investigated to examine whether class imbalance can be resolved with our research.

\end{enumerate}

% --------------------------------

% Part 6
% --------------------------------
\section{Experimental Setup and Result}

The experiments conducted in this study can be categorized in 2 approaches. First, the mutual information approach is conducted on a smaller dataset to avoid concerns related to computational resources. On the other hand, the support points approach is performed on a larger dataset to highlight the potential in how our research can be applied in large datasets.

\subsection{Mutual Information Approach}

\subsubsection{Experimental Setup}
The mutual information approach aimed to thoughtfully group similar data together in a stratum, so when simple random sampling is conducted afterwards, it can sample on strata that may have larger between variances with other strata. This approach is examined on an imbalance breast cancer dataset where most samples are not diagnosed with breast cancer. By conducting this experiment, we can observe the impact of the undersampling methodology proposed, along with how well the undersampled data represent the original data. 

% The \textbf{mutual information} approach involves two major steps:
% \begin{enumerate}
%     \item \textbf{Stage 1}: Computing pairwise mutual information for the majority class samples in the dataset.
%     \item \textbf{Stage 2}: Determine the optimal number of strata by converting the mutual information values into dissimilar distances, then perform clustering with the elbow method.
%     \item \textbf{Stage 3}: Perform simple random sampling on the majority class data with the stratum determined earlier, and using either Neyman or Optimal allocation.
%     \item \textbf{Stage 4}: Combine the undersampled majority class samples with the minority class samples for further classification work.
% \end{enumerate}

The dataset used contains 272 samples with 15 features, where the majority class (patients who do not have breast cancer) accounts for approximately 70.22\% of the data, and the minority class (patients who has breast cancer) makes up the remaining 29.78\%. The entire optimization process for mutual information required less than \textbf{1 minute}. This is because the dataset fitted is very small. We expect the computation of pairwise MI will increase significantly with a larger dataset.

The mutual information approach follows a structured four-stage process as detailed below:
\begin{enumerate}
\item \textbf{Stage 1: Pairwise Mutual Information Computation}

The mutual information is computed among all majority samples to observe how similar each one is compared to each other. The computation method can be found in section 4.1.2.

\item \textbf{Stage 2: Determine the Optimal Number of Strata}

We use the elbow method to determine the optimal number of strata. Specifically, we converted the mutual information values from the mutual information matrix to a dissimilarity distance matrix for future clustering work. By clustering the majority class data based on their dissimilarity distances, we strive to find the optimal number of strata that has both low within-cluster variance as well as avoid over-partitioning the dataset. Therefore the elbow method becomes the optimal choice. 
In our experiment, the optimal strata number is 4 so we will use that number for the rest of the experiment in this approach.

\item \textbf{Stage 3: Perform Stratified SRS on Majority Class Data}

Now given the optimal strata number is 4, we performed stratified SRS on the majority class data by sampling a specific amount of samples in each stratum. After performing SRS once, if the number of data sampled did not reach to the size of the minority class because some strata has been fully sampled, we will conduct SRS repeatedly on other strata with samples until the number of data sampled reached to the size of the minority class.

As mentioned above, the specific amount of samples sampled in each stratum is determined by an allocation strategy utilized in stratified SRS. This experiment explore Neyman and Optimal allocation strategies. They are fairly similar as their main differences comes from the additional cost consideration in optimal allocation. In other words, Neyman allocation can be viewed as a special case of optimal allocation where the cost function is uniform. For our experiment, the cost function is modified in a way that the costs are higher for larger size strata. We hope this can incentivize the sampling process to sample data from smaller strata and avoid sampling too many data from the large strata.

\item \textbf{Stage 4: Combine undersampled majority class with minority class data}

Lastly, the resampled data can be formed by simply merging the undersampled majority class data with the unvaried minority class data. We also reshuffled the data to avoid any potential bias when the resampled data is fitted to the classifier later. 
\end{enumerate}

\subsubsection{Modeling and Comparison Against Naive Downsampling}
In this experiment, we trained four classifiers: Logistic Regression, Random Forest, Extreme Gradient Boosting (XGBoost), and Support Vector Machine (SVC) on the breast cancer using two different downsampling techniques: Random Downsampling as baseline; and mutual information as treatment. 

The models were evaluated using metrics suited for imbalanced data: Balanced Accuracy, Precision, Recall, and F1-score. The results showed that mutual information consistently outperformed or matched random downsampling across all models. Specifically, there is a significant improvement when the resampled data fitted to the logistic regression, random forest, and support vector machine with over 30\%, 20\%, and 12\% respectively.

% For instance:
% \begin{itemize}
%     \item \textbf{Logistic Regression}: Balanced Accuracy = 0.8485 (Mutual Information) vs 0.5455 (Naive).
%     \item \textbf{Random Forest}: Balanced Accuracy = 0.6970 (Mutual Information) vs 0.4848 (Naive).
%     \item \textbf{XgBoost}: Balanced Accuracy = 0.5455 (Mutual Information) vs 0.5152 (Naive).
%     \item \textbf{SVC}: Balanced Accuracy = 0.7273 (Mutual Information) vs 0.6061 (Naive).
% \end{itemize}

% Table: Accuracy
\begin{table}[ht]
\centering
\small
\caption{Summary of Classification Accuracies for each Combination of Classifier Model and Undersampling Method.}
\label{tab:feature-means}
\begin{tabular}{|c|c|c|}
\hline
\textbf{Classifier Model} & \textbf{Undersampling Method} & \textbf{Accuracy} \\ \hline
Logistic Regression  & Random Undersampling & 0.5455 \\ \hline
Logistic Regression & Mutual Information & 0.8485 \\ \hline
Random Forest & Random Undersampling & 0.4848  \\ \hline
Random Forest & Mutual Information & 0.6970  \\ \hline
XGBoost & Random Undersampling & 0.5152  \\ \hline
XGBoost & Mutual Information & 0.5455  \\ \hline
SVC & Random Undersampling & 0.6061  \\ \hline
SVC & Mutual Information & 0.7273  \\ \hline
\end{tabular}
\end{table}

% \subsubsection{Computational Aspects}
% The entire optimization process for mutual information required less than \textbf{1 minute}. This is because the dataset fitted is very small. We expect the computation of pairwise MI will increase significantly with a larger dataset.

\subsubsection{Summary of Results}
Overall, this approach works efficiently in a small dataset when the MI computation is relatively lightweight. The classification performances from the undersampled data when adopted this approach have outperformed this experiment's baseline: random under sampling, achieving significantly higher balanced accuracy, recall, and f1-scores. Altogether, this approach suggests our study may bring improvements in representativeness when undersampling majority class data.

\subsection{Support Points Approach}

\subsubsection{Experimental Setup}
This experiment addresses the severe class imbalance in a credit card fraud detection dataset using the \textit{Support Points} approach. The dataset contains 284,807 samples with 30 features, where the majority class (non-fraudulent transactions) accounts for 99.82\%, and the minority class (fraudulent transactions) constitutes only 0.18\%. The primary objective is to select a representative subset of the majority class while preserving its statistical properties, thus avoiding the pitfalls of naive random downsampling.

Given the dataset's size, direct computation of pairwise distances for support point optimization is computationally prohibitive, requiring over 600GB of memory. To overcome this, we employ a \textbf{two-stage approach}:

\begin{enumerate}
    \item \textbf{Stage 1: Clustering-Based Subsampling}
    
    The majority class is clustered into 50 groups using MiniBatchKMeans, with the number of clusters determined based on the Elbow Method. This ensures a balance between granularity and computational efficiency. From these clusters, a proportional subset of 5,000 samples is extracted to represent the majority class while retaining its heterogeneity.
    
    \item \textbf{Stage 2: Generating Support Points}

    Using the reduced majority subset, we generate support points by optimizing the selected samples to minimize the energy distance between the reduced majority class and the original majority class. A gradient-based optimization algorithm iteratively refines the support points over \textbf{2000 iterations}, achieving a final energy distance of approximately \textbf{0.0102}. This ensures that the generated support points maintain statistical fidelity to the original data while significantly reducing its size.
\end{enumerate}

This experimental setup ensures that the selected subset aligns closely with the original majority distribution while addressing computational limitations, enabling effective training for imbalanced classification tasks.

\subsubsection{Quantitative Validation of Representativeness}

To validate the representativeness of the downsampled subsets, we conducted a \textit{Feature-wise Statistics} analysis and a \textit{Kolmogorov-Smirnov (KS) Test}, comparing the original majority class, naive downsampling, and the support points subset (\(Z_{\text{nearest}}\)).

\begin{enumerate} \item \textbf{Feature-wise Statistics Analysis}

The feature-wise comparison of means and standard deviations revealed that both the Support Points and Naive Downsampling methods closely approximate the original majority class. A summary of feature-wise statistics is shown below:

\begin{figure}[ht]
    \centering
    \includegraphics[width=\linewidth, keepaspectratio]{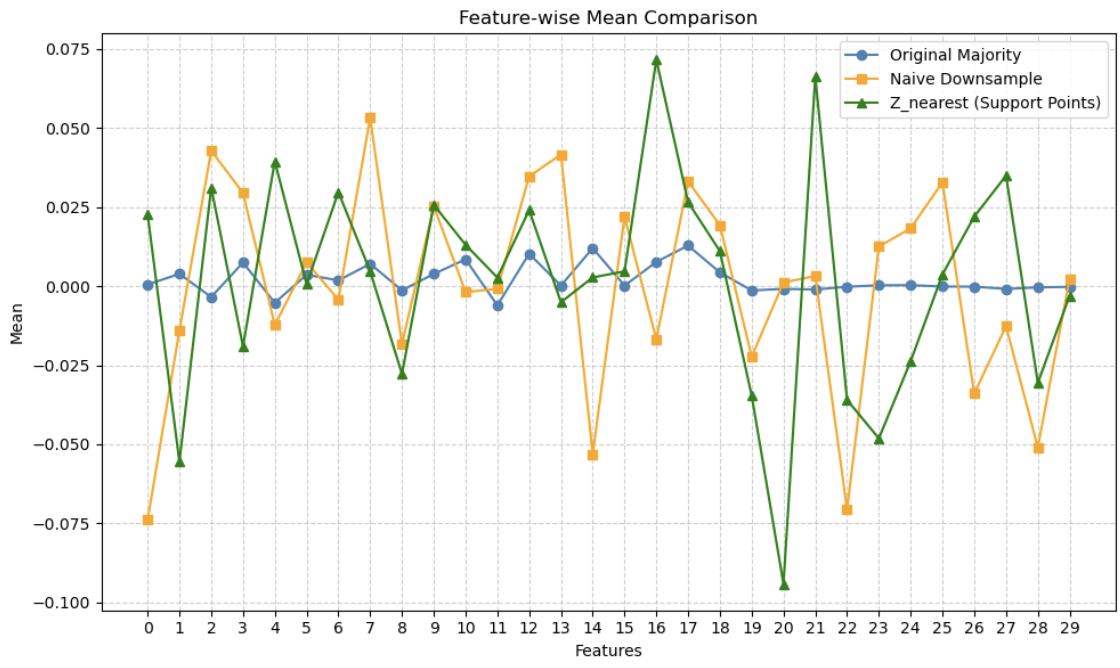}
    \caption{Feature-wise Mean Comparison for Support Points and Naive Downsampling.}
    \label{fig:feature-wise-mean-comparison}
\end{figure}

\begin{figure}[ht]
    \centering
    \includegraphics[width=\linewidth, keepaspectratio]{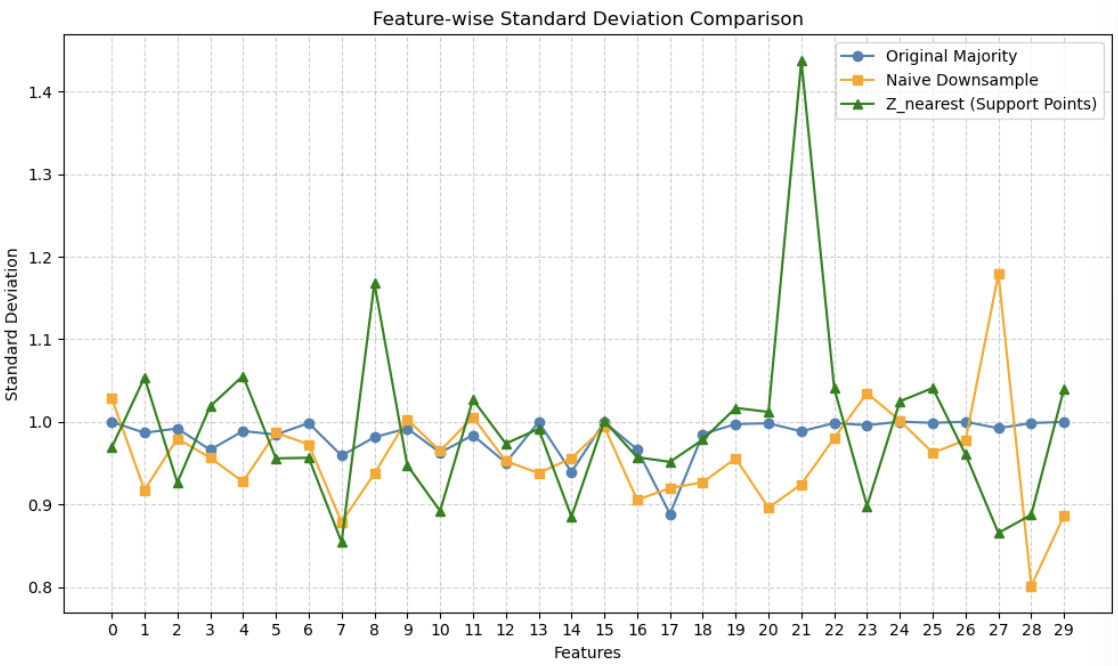}
    \caption{Standard Deviation Comparison for Support Points and Naive Downsampling.}
    \label{fig:standard-deviation-comparison}
\end{figure}

% First Table: Means
\begin{table}[ht]
\centering
\small
\caption{Feature-wise Means for Support Points and Naive Downsampling.}
\label{tab:feature-means}
\begin{tabular}{|c|c|c|c|}
\hline
\textbf{Feature} & \textbf{Original Mean} & \textbf{\(Z_{\text{nearest}}\) Mean} & \textbf{Naive Mean} \\ \hline
Feature 5  & 0.0036 & 0.0006 & 0.0075 \\ \hline
Feature 12 & 0.0102 & 0.0240 & 0.0347 \\ \hline
Feature 21 & -0.0011 & 0.0663 & 0.0032 \\ \hline
\end{tabular}
\end{table}

% Second Table: Standard Deviations
\begin{table}[ht]
\centering
\small
\caption{Feature-wise Standard Deviations for Support Points and Naive Downsampling.}
\label{tab:feature-std}
\begin{tabular}{|c|c|c|c|}
\hline
\textbf{Feature} & \textbf{Original Std} & \textbf{\(Z_{\text{nearest}}\) Std} & \textbf{Naive Std} \\ \hline
Feature 5  & 0.9846 & 0.9558 & 0.9870 \\ \hline
Feature 12 & 0.9504 & 0.9737 & 0.9525 \\ \hline
Feature 21 & 0.9886 & 1.4370 & 0.9244 \\ \hline
\end{tabular}
\end{table}

The Feature-wise Mean Comparison and Standard Deviation Comparison plots highlight these trends, showing that both approaches align well with the original data distribution. While some features exhibited slight deviations, particularly in standard deviation (e.g., Feature 21), the overall fidelity to the original distribution is notable.

\item \textbf {Kolmogorov-Smirnov (KS) Test Results}

The KS test results further validate the representativeness of the downsampled subsets. For most features, both methods yielded high p-values, indicating no statistically significant differences from the original majority class. A summary of selected KS test results is provided below:

\begin{table}[ht]
\centering
\small
\caption{Selected KS Test Results for Support Points and Naive Downsampling.}
\label{tab:ks-test}
\begin{tabular}{|c|c|c|c|c|}
\hline
\textbf{Feature} & \textbf{KS(Z)} & \textbf{P(Z)} & \textbf{KS(Naive)} & \textbf{P(Naive)} \\ \hline
Feature 8  & 0.0232 & 0.9556 & 0.0283 & 0.8331 \\ \hline
Feature 14 & 0.0407 & 0.4055 & 0.0650 & 0.0352 \\ \hline
Feature 24 & 0.0208 & 0.9846 & 0.0327 & 0.6809 \\ \hline
\end{tabular}
\end{table}

These results indicate that both methods maintain the overall distributional properties of the original data, with Support Points exhibiting slightly better fidelity in specific features. However, the observed differences are minor, and the methods perform comparably under these metrics.
\end{enumerate}

\subsubsection{Modeling and Comparison Against Naive Downsampling}
In this experiment, we trained four classifiers—Logistic Regression, Random Forest, XGBoost, and SVC—on datasets created using two different downsampling techniques: Support Points and Naive Random Downsampling. The models were evaluated using metrics suited for imbalanced datasets: Balanced Accuracy, Precision, Recall, and F1-score.

% Table: Accuracy
\begin{table}[ht]
\centering
\small
\caption{Summary of Classification Accuracies for each Combination of Classifier Model and Undersampling Method.}
\label{tab:feature-means}
\begin{tabular}{|c|c|c|}
\hline
\textbf{Classifier Model} & \textbf{Undersampling Method} & \textbf{Accuracy} \\ \hline
Logistic Regression  & Random Undersampling & 0.9261 \\ \hline
Logistic Regression & Support Points & 0.9401 \\ \hline
Random Forest & Random Undersampling & 0.9295  \\ \hline
Random Forest & Support Points & 0.9330  \\ \hline
XGBoost & Random Undersampling & 0.9225  \\ \hline
XGBoost & Support Points & 0.9225  \\ \hline
SVC & Random Undersampling & 0.9260  \\ \hline
SVC & Support Points & 0.9331  \\ \hline
\end{tabular}
\end{table}

Across all models, the Support Points approach consistently performed slightly better or matched the performance of Naive Random Downsampling. For instance, Logistic Regression achieved a Balanced Accuracy of 0.94 with Support Points compared to 0.926 with Naive Downsampling, demonstrating the effectiveness of the Support Points method in maintaining data representativeness while addressing class imbalance.

Overall, the support points approach provides a principled and effective method for addressing class imbalance while preserving the distributional integrity of the majority class.
% --------------------------------

% Part 7
% --------------------------------
\section{Contribution and Significance}

\subsection{Mutual Information}

 The current research about mutual information emphasizes on brainstorming efficient and accurate estimates about mutual information. Our work explores the potential application of mutual information along with clustering algorithms in the context of data resampling. The concept of mutual information allows stratum to better group similar data points, which allows the sampling algorithm to select a variety of representative samples, thereby retaining dataset distillation.

 Compared to other undersampling techniques where datapoints are removed either randomly, or based their decision boundaries, or their neighboring datapoints, the mutual information approach provided a new metric for researchers to consider when they are looking for criteria to select more representative datapoints. This approach can be reapplied to many classification tasks in real life with limited dataset, which is a common phenomenon in industries where data is hard to obtain, especially medical and government institutions.

\subsection{Support Points}

Our work introduces a novel application of \textit{support points} for downsampling imbalanced datasets in classification tasks, inspired by the method developed by Mak and Joseph (2018)\cite{Mak-Support-Points}. Support points, which minimize the \textit{energy distance} to preserve the statistical properties of the original distribution, have primarily been used in uncertainty quantification and Bayesian computation. Here, we extend their use to class imbalance problems, providing a principled approach to select representative subsets from the majority class while preserving the integrity of the original data distribution. 

Our contribution highlights the adaptability of support points to machine learning challenges, offering a new perspective on downsampling methods for tackling class imbalance issues effectively. Unlike traditional random downsampling methods, which may compromise statistical properties or discard critical data, support points retain representative diversity and structure within the dataset. This ensures improved performance for machine learning models in high-stakes domains such as medical diagnosis, fraud detection, and rare event prediction. 

By bridging the gap between statistical sampling and machine learning, this work opens up new opportunities to address imbalanced learning challenges with greater fairness and robustness.

% --------------------------------

% Part 8
% --------------------------------
\section{Conclusion and Future Work}

\subsection{Support Points}

Our work demonstrates the effectiveness of \textit{support points} as a statistically sound and computationally efficient downsampling technique for addressing class imbalance. By optimizing the selection of representative majority samples, support points preserve the statistical properties of the original dataset better than traditional random downsampling methods. This approach reduces data redundancy while maintaining feature-wise integrity, contributing to improved model performance in imbalanced classification tasks. However, the computational cost of generating support points remains a notable limitation, especially for large datasets. Additionally, while the method retains global patterns effectively, it may miss finer, localized data nuances, which could impact specific applications.

Future efforts will focus on enhancing the computational efficiency of support point generation, leveraging advanced optimization techniques such as accelerated gradient methods and parallelized algorithms. Extending the method to handle extremely large datasets in a scalable manner will be critical for real-world applicability. Furthermore, exploring the adaptability of support points to high-dimensional, heterogeneous data structures, including time series and graph-based datasets, presents a promising research direction. Finally, validating this approach across diverse, domain-specific applications will solidify its practical impact and uncover opportunities for further refinement and optimization.

\subsection{Mutual Information}

In conclusion, utilizing mutual information as a metric to measure how similar datapoints are for further stratification and sampling has shown to be an effective undersampling approach when the dataset is small. This suggests that mutual information may help retain the representativeness of the sampled data, further promoting the distillation of the dataset. 

However, this study did not investigate the application of computing pairwise mutual information in a larger dataset. As the brute force computing approach mentioned earlier foreshadowed the potential difficulties of such implementation. As a result, this study opened more research questions about the efficiency and generalizability of this approach, including the major concern on how can this approach be still robust and efficient in a large dataset.  We hope our study can inspire more upcoming research in this area, which may eventually lead to more representative undersampling techniques to address class imbalance in machine learning tasks. 

% --------------------------------

% DONE STUFF

% % Acknowledgements should only appear in the accepted version.
% \section*{Acknowledgements}

% \textbf{Do not} include acknowledgements in the initial version of
% the paper submitted for blind review.

% If a paper is accepted, the final camera-ready version can (and
% probably should) include acknowledgements. In this case, please
% place such acknowledgements in an unnumbered section at the
% end of the paper. Typically, this will include thanks to reviewers
% who gave useful comments, to colleagues who contributed to the ideas,
% and to funding agencies and corporate sponsors that provided financial


\bibliography{main.bib}

\begin{thebibliography}{6}
\providecommand{\natexlab}[1]{#1}
\providecommand{\url}[1]{\texttt{#1}}
\expandafter\ifx\csname urlstyle\endcsname\relax
  \providecommand{\doi}[1]{doi: #1}\else
  \providecommand{\doi}{doi: \begingroup \urlstyle{rm}\Url}\fi

\bibitem[Devi et~al.(2020)Devi, Biswas, and Purkayastha]{Devi}
Devi, D., Biswas, S.~K., and Purkayastha, B.
\newblock A review on solution to class imbalance problem: Undersampling
  approaches.
\newblock In \emph{2020 International Conference on Computational Performance
  Evaluation (ComPE)}, pp.\  626--631, 2020.
\newblock \doi{10.1109/ComPE49325.2020.9200087}.

\bibitem[Ferdosi et~al.(2020)Ferdosi, Gholamidavoodi, and
  Mohimani]{pmlr-v108-ferdosi20a}
Ferdosi, M., Gholamidavoodi, A., and Mohimani, H.
\newblock Measuring mutual information between all pairs of variables in
  subquadratic complexity.
\newblock In Chiappa, S. and Calandra, R. (eds.), \emph{Proceedings of the
  Twenty Third International Conference on Artificial Intelligence and
  Statistics}, volume 108 of \emph{Proceedings of Machine Learning Research},
  pp.\  4399--4409. PMLR, 26--28 Aug 2020.
\newblock URL \url{https://proceedings.mlr.press/v108/ferdosi20a.html}.

\bibitem[Fernandes \& Gloor(2010)Fernandes and Gloor]{Fernandes2010}
Fernandes, A.~D. and Gloor, G.~B.
\newblock Mutual information is critically dependent on prior assumptions:
  would the correct estimate of mutual information please identify itself?
\newblock \emph{Bioinformatics}, 26\penalty0 (9):\penalty0 1135--1139, 2010.
\newblock \doi{10.1093/bioinformatics/btq111}.
\newblock URL \url{https://doi.org/10.1093/bioinformatics/btq111}.

\bibitem[Ganganwar(2012)]{Ganganwar}
Ganganwar, V.
\newblock An overview of classification algorithms for imbalanced datasets.
\newblock \emph{International Journal of Emerging Technology and Advanced
  Engineering}, 2:\penalty0 42--47, 01 2012.

\bibitem[Mak \& Joseph(2018)Mak and Joseph]{Mak-Support-Points}
Mak, S. and Joseph, V.~R.
\newblock Support points.
\newblock \emph{The Annals of Statistics}, 46\penalty0 (6A):\penalty0
  2562--2592, 2018.
\newblock ISSN 00905364, 21688966.
\newblock URL \url{https://www.jstor.org/stable/26542875}.

\bibitem[Shang et~al.(2024)Shang, Yuan, and Yan]{NEURIPS2023_24d36eee}
Shang, Y., Yuan, Z., and Yan, Y.
\newblock Mim4dd: mutual information maximization for dataset distillation.
\newblock In \emph{Proceedings of the 37th International Conference on Neural
  Information Processing Systems}, NIPS '23, Red Hook, NY, USA, 2024. Curran
  Associates Inc.

\end{thebibliography}
\bibliographystyle{icml2021}

\end{document}